\pdfoutput=1
\documentclass[11pt]{article}

\usepackage[preprint]{acl}

\usepackage{times}
\usepackage{latexsym}
\usepackage{subcaption}
\usepackage{amsmath}
\usepackage{enumitem}

\usepackage{tabularray}

\usepackage[T1]{fontenc}
\usepackage[utf8]{inputenc}

\usepackage{microtype}

\usepackage{inconsolata}

\usepackage{graphicx}

\DeclareMathAlphabet{\mathcal}{OMS}{cmsy}{m}{n}

\title{Unsupervised Location Mapping for Narrative Corpora}

\author{Eitan Wagner$^\dagger$\quad Renana Keydar$^\ddagger$ \quad Omri Abend$^\dagger$ \\
$^\dagger$ Department of Computer Science \quad
         $^\ddagger$ Faculty of Law and Digital Humanities\\
         Hebrew University of Jerusalem\\ \texttt{\{first\_name\}.\{last\_name\}@mail.huji.ac.il}}
\begin{document}
\maketitle

\begin{abstract}
This work presents the task of {\it unsupervised location mapping}, which seeks to map the trajectory of an individual narrative on a spatial map of locations in which a large set of narratives take place. 
Despite the fundamentality and generality of the task, very little work addressed the spatial mapping of narrative texts.
The task consists of two parts: (1) inducing a ``map'' with the locations mentioned in a set of texts, and (2) extracting a trajectory from a single narrative and positioning it on the map.
Following recent advances in increasing the context length of large language models, we propose a pipeline for this task in a completely unsupervised manner without predefining the set of labels.
We test our method on two different domains: 
(1) Holocaust testimonies and (2) Lake District writing, namely multi-century literature on travels in the English Lake District. 
We perform both intrinsic and extrinsic evaluations for the task, with encouraging results, thereby setting a benchmark and evaluation practices for the task, as well as highlighting challenges.\footnote{Our codebase will be released upon publication.}
\end{abstract}

\section{Introduction}
The grounding of events in locations is often seen as a defining characteristic that sets narrative texts apart from other types of writing \citep{piper2022toward}. Thus, the trajectory of a narrative, i.e., the sequence of locations in which it takes place, is an essential aspect. 
Characterizing a story by a sequence of locations is also beneficial as a backbone for alignment between different stories -- an important task in its own right \citep[see, e.g.,][]{ernst-etal-2022-proposition}. 
Additionally, as we will see, location extraction is a task that requires long-range narrative understanding, a highly active topic in NLP \cite{yao-etal-2022-corpus, bertsch2024unlimiformer}.

However, despite the abundance of Natural Language Processing (NLP) research on identifying locations in texts, few efforts have been made to extract the progression or sequence of locations from a narrative story \citep{wagner-etal-2023-event}. As a structured prediction task with a large class set, the ability to obtain sufficient data for generalization is very limited.

In this work, we present the task of zero-shot trajectory mapping and design a pipeline for it with long-context large language models. 
Zero-shot trajectory mapping involves both the extraction of the locations for each document (as a ``trajectory'') and the identification of the relationship between the locations (creating a ``map''). The task assumes no predefined set of locations but rather seeks to construct a map based only on the given texts. 
Thus, the task is unsupervised in two senses -- the set of locations must be inferred from a set of unannotated texts, and the trajectory of each text must be extracted without supervision.

\begin{figure*}[t]
\centering
\includegraphics[trim={0 0 0 0}, scale=0.15]{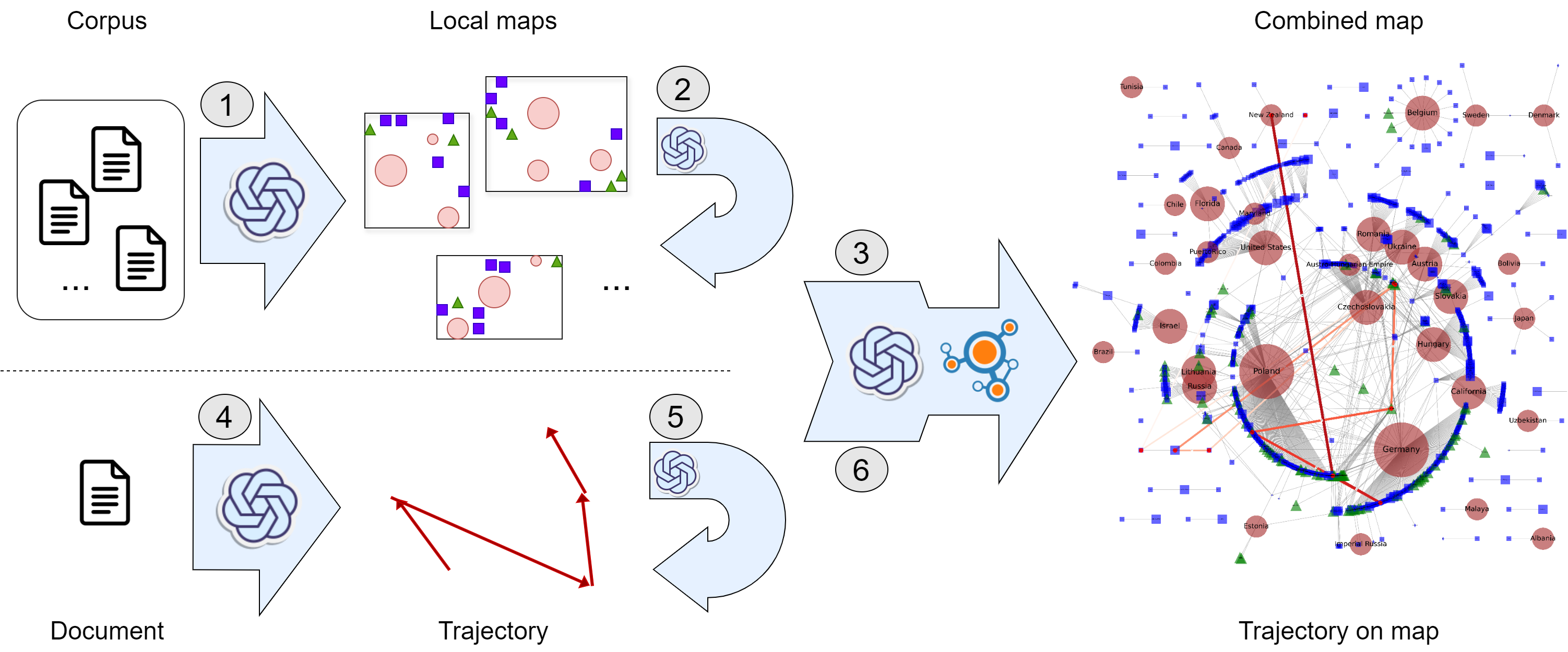}
\caption{Overview of the pipeline. The top path represents the creation of a combined map, with steps: (1) per-document location graph extraction, (2) revision, and (3) combining the graphs and visualizing. The bottom path represents mapping a trajectory, with steps: (4) per-document trajectory extraction, (5) revision, and (6) mapping on the combined graph.}
\label{fig: overview}
\end{figure*}

We experiment on two corpora: 
(1) Holocaust survivor testimonies, and 
(2) works describing the English Lake District \citep{10.1145/3149858.3149865}. 
We select these corpora as they both include a variety of spatial descriptions in a relatively confined geographical setting. This sets them apart from typical narrative datasets, which are either limited in the number of documents or unrestricted in the possible locations \citep{sultananarrative}.
While the documents in each corpus are confined, each corpus has its own distinct setting, in terms of the set of places (i.e., in what countries they are) and their physical size and specificity (e.g., from countries and cities to castles and lakes).\footnote{We note that the corpora are significantly different in their sensitivity and method of collection. Each corpus is unique and is worthy of individual research. Our work demonstrates how a general pipeline can be applied to various domains.}

We design a pipeline for zero-shot trajectory mapping and implement it using GPT-4o mini.
An overview of the pipeline is displayed in Figure \ref{fig: overview}.
We apply the method to $402$ testimonies and $75$ Lake District works.
We evaluate the resulting maps and discuss potential uses, such as alignment between testimonies. Our task serves as another challenging test-bed for long-context LLMs in the context of narrative understanding and our work demonstrates the effectiveness of these models. 

To recap, the contributions in this work are:
(1) formally presenting a new task of (unsupervised) trajectory mapping;
(2) proposing a simple method for the task that leverages long-context LLMs;
(3) demonstrating the efficacy of the pipeline, both intrinsically and extrinsically, on diverse domains;
(4) discussing theoretical and practical challenges that arise from the unsupervised nature of the task.

%%%%%%%%%%%%%%%%%%%%%%%%%%%%%%%%%%%%%%%%%%%%%%%%%%
\section{Previous Work} \label{sec:previous-work}

\paragraph{Narrative Analysis.}
Narrative schema analysis aims to capture the core of event sequences, providing a condensed sequential timeline of a lengthy story. This overview helps in aligning relevant parts and identifying common topic paths, as demonstrated by \citet{antoniak2019narrative} in their study on birth stories using segment-wise topic modeling.

To extract an interpretable sequential progression it was assumed necessary to divide the long story into shorter segments \citep{wagner-etal-2023-event}. However, recent advances in NLP introduced significant increases in context lengths of models \cite{wang2024limits}, allowing the extraction of sequences as an end-to-end task.

Recent studies have highlighted the importance of event locations in narrative analysis. \citet{piper-etal-2021-narrative} provided a definition of narratives that included a focus on event locations. \citet{soni2023grounding} introduced a task involving grounding characters in specific locations. \citet{KUMAR2019365} extracted event locations from individual events, such as those found in tweets. \citet{wagner-etal-2023-event} expanded on this concept by examining trajectories of locations throughout entire narratives, using a predetermined set of coarse-grained categories. \citet{wilkens2024small} investigated the mobility of characters in fictional and non-fictional narratives. 

\paragraph{Trajectory Modeling in Transportation.} 
Another line of work extracts document-level trajectories in transportation.
\citet{wesley2012predicting} applied Hidden Markov Models (HMM) to human location trajectories. \citet{sassi2019location} used convolutional neural networks on location embeddings as an alternative to HMMs. \citet{lui-etal-2021} employed LSTM-based models for predicting pedestrian trajectories.
These works focus on locations given as coordinates and not as natural text descriptions, which allow for a more thematic level of representation and comparison \cite{wagner-etal-2023-event}.

\paragraph{Narrative Cartography.}
Many works investigated the mapping of narratives. \citet{reuschel2011mapping} presented methods for the visualization of location maps. Their methods show differences between the maps in fiction and non-fiction. \citet{mai2022narrative} developed toolboxes for enrichment of geographic data, based on knowledge graphs. 

These works are primarily based on a location ontology, thus limiting the scope to domains with sufficient prior knowledge. In our work, we propose a completely unsupervised method, allowing its application without any prior knowledge.

%%%%%%%%%%%%%%%%%%%%%%%%%%%%%%%%%%%%%%%%%%%%%%%%%%
\section{Trajectory Mapping}

\subsection{Task Definition}

We are given a set of texts $\mathbf{x}^1, \mathbf{x}^2, \ldots  \mathbf{x}^k$, each divided into sentences, $\mathbf{x}^i =x^i_1, x^i_2, ..., x^i_n$.\footnote{This definition is agnostic to the actual segmentation method. In our experiments we used sentences, but we can also use larger or smaller segments.}
The task aims to produce two outputs: 

\begin{enumerate}
    \item 
        {\bf A map:} a directed graph $G=(V, E)$, where the vertices $V$ are all the locations (name+type) in the set of texts, and the edges $E$ are the relationships between them (e.g., New York is in the United States).
    \item 
        {\bf Trajectories:} for each $\mathbf{x}^i$, a path $(v^i_1, v^i_2, \ldots, v^i_k) \in V^k$ that reflects the trajectory (i.e., the sequence of locations in which the events take place) in this text. 
        We require adjacent vertices to be different but allow non-adjacent repetition.
        The path should have additional vertex labels for the indices within the text of this location (e.g., segments 17-21) and edge labels for the method of transportation, if applicable (e.g., ``by foot'', ``by plane'' etc.). 
\end{enumerate}

It is instructive to compare both parts of the task to traditional Named Entity Recognition (NER) for location categories. NER is a phrase-level prediction task that ignores the relationship between different locations or even between mentions of the same locations. Therefore, the first part of our task can be seen as a combination of NER and Entity Relation Extraction (focusing on the containment relation). 
The second part of our task is completely different as it requires a structured sequence as an output. Specifically, the trajectory describes a sequence of transitions and not just isolated mentions. 
For example, if the text introduces a person who came from some named place, NER should mark the mention, while location mapping will not include it as part of the trajectory.
 
We also remark that the second task differs from supervised location tracking \cite{wagner-etal-2023-event} in two respects: (1) the task is not limited by granularity -- it extracts countries, cities and other types of locations (e.g., ``the forest''); (2) the task considers only locations that are mentioned in the text. This is also a challenge since texts might differ in their tendency to mention locations, leading to different outputs for the same trajectory.

\subsection{Evaluation}

\paragraph{Evaluating Maps.} \label{map_eval}
Since full comparison between graphs can be noisy, in our evaluation we only compare edges between graphs with the same node set. We report accuracy metrics -- precision, recall, and F1-score.

Formally, we define $\text{TP}=|E_m \cap E_r|$, $\text{FP}=|E_m \cap E_r^c|$, and $\text{FN}=|E_m^c \cap E_r|$, where $E_m$ and $E_r$ are the edges from the model and the reference, respectively. Then our metrics are
$$ \text{precision} = \frac{TP}{TP+FP}, \quad \text{recall} = \frac{TP}{TP+FN}$$

and
$$ \text{F1} = 2 \cdot \frac{\text{precision} \cdot \text{recall}}{\text{precision} + \text{recall}}$$

\paragraph{Evaluating Trajectories.} \label{trajectory_eval}
Given a reference trajectory, we want to compare the model's output to the reference.
A natural choice would be their edit distance \cite{10.1145/363958.363994}, which counts the minimal number of edits between the sequences.

A difficulty in our case is that the trajectory length is not fixed. This means that the reference trajectories of different lengths have different impact on the aggregated score. To alleviate this, we normalize by the length of the reference sequence. We denote this by {\sc Edit}.\footnote{This normalized distance, is known in the literature as {\it word error rate}. It has some undesired properties as it might be larger than $1$ and is not a proper distance function. It does, however, allow us to gain an insight into the performance over multiple cases.} 

Another challenge regarding lengths is that there might also be systematic differences in lengths due to different interpretations of the task. For example, one source might mark a room as a location while another will not. Unnormalized edit distance might thus be unjustly biased towards outputs with lengths closer to the reference.
Therefore, we additionally report a modified version of the edit distance that is recall-oriented. In this version, we give no penalty for the deletion of locations in the predicted document. We denote this measure with ({\sc R-Edit}).

\section{Data}

\subsection{Holocaust Testimonies}

Our dataset consists of $1000$ Holocaust survivor testimonies, received from the Shoah Foundation (SF).\footnote{\url{https://sfi.usc.edu/}} All interviews were conducted face-to-face by an interviewer, recorded on video, and transcribed as time-stamped text. The lengths of the testimonies range from $2609$ to $88105$ words, with a mean length of $23536$ words. 

\paragraph{Reference Data.}\label{sec:eval}
For evaluation, we use the test set in \citet{wagner-etal-2023-event}, originally constructed for supervised location tracking. 
This test set is based on the SF annotations, which are highly detailed tags given to one-minute segments. The annotations were completed and proofed by domain experts to create trajectories.
Since the SF labels were given to relatively large segments, they are limited as annotations for zero-shot trajectory extraction which can be more detailed. Since we expect labels in the annotation to appear in any zero-shot extraction, we use this test set to compute a recall-focused metric.

Additionally, we re-annotated testimonies by two annotators, with the same instructions given to the language models. One annotator did $6$ testimonies and the other did $3$ (out of the $6$). This annotation included multiple revisions with detailed guidelines regarding what locations should be included.
We denote these sets by {\sc Ref1} and {\sc Ref2}. We denote the set of SF annotations on these testimonies by {\sc SF-Ref}. The annotation guidelines are identical to the LLM prompts (\S \ref{prompt1}, \S\ref{prompt2}) up to formatting constraints.
Despite the effort to create unambiguous guidelines, the difference in the number of locations between the annotators was very high, hence these reference documents cannot be regarded as the only possible outputs.

Altogether, for the evaluation of the trajectory task (task 2), we have two reference annotations on $6$ documents and another reference for $3$ of the documents.
We do not have annotations for the location map (task 1) in this dataset.

\subsection{Corpus of Lake District Writing}

The Corpus of Lake District Writing \citep[CLDW;][]{10.1145/3149858.3149865}, consists of $80$ annotated texts about the English Lake District. 
The texts belong to various genres, such as travel journals, novels, and poetry, and have a large date range, from 1622 to 1900.
The length varies from $1063$ words to $95523$ words, with a mean of $19022$.

\paragraph{Reference data.}\label{sec:eval2}
The CLDW has annotations for named entities with their geographic coordinates (GIS labels). 
These annotations do not necessarily define a trajectory (task 2), which we define as the sequence of locations in which the recounted events take place. For example, if place A is described as far from place B, A will also be marked as a named entity, but should not be included in the trajectory.

For the mapping task (1), we can use the GIS labels to create an approximate map for a given set of locations. We divide the locations into levels (Country, County, City, Natural, and Facility) and create a hierarchical tree based on proximity. That is, we connect each natural location and facility to the nearest city and each city to the nearest county. We can then compare the output of a suggested model to this graph using standard metrics (such as the F1 score).\footnote{We use only the set of predicted locations and not the full set of GIS labels since we do not require all named locations to be in the map. Therefore, this metric can be seen as focused on precision and not recall.}

\begin{figure*}[t]
\centering
\includegraphics[scale=0.118]{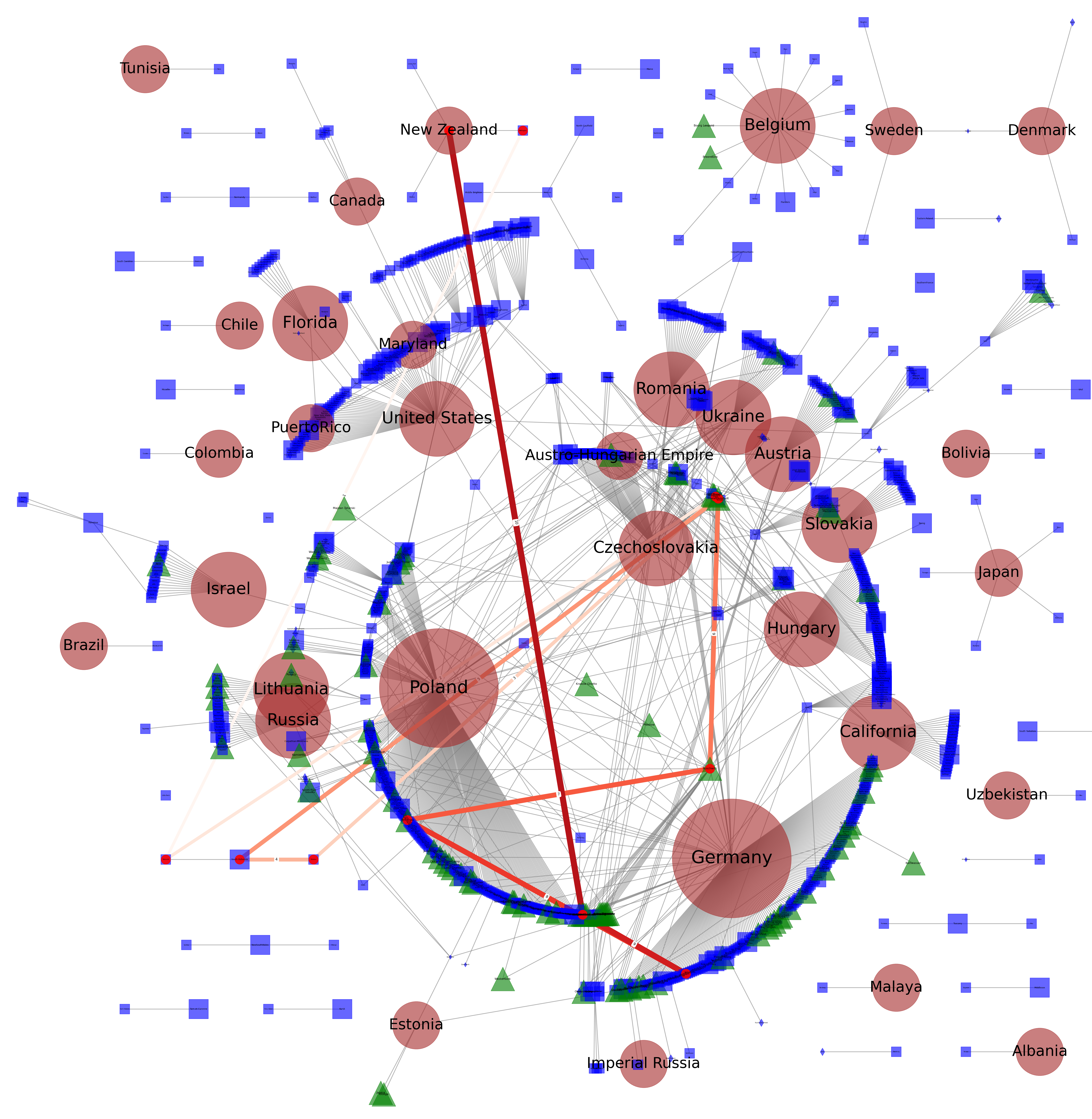}
\caption{Visualization of a location map with a single trajectory. The map was generated based on 402 testimonies with GPT-4o-mini. Some low-degree nodes were removed for clarity. Countries are displayed as brown circles, where the size depends on the degree. Cities are blue squares. Holocaust-related locations are green triangles. The trajectory is in shades of red, getting darker with the progression of the trajectory.} \label{fig: figure1}
\end{figure*}

\begin{figure}[t]
\centering
\includegraphics[scale=0.66]{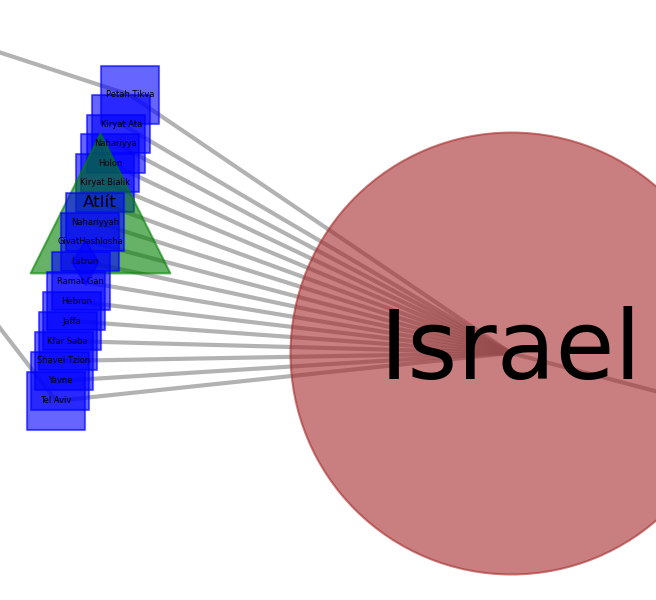}
\caption{Snippet from the map that includes Israel and locations within it.} \label{fig: figure2}
\end{figure}

\section{Zero-shot Trajectory Mapping Pipeline}

Recent advances in LLMs led to a substantial increase in the context window that serves as input to the models.\footnote{\url{https://www.anthropic.com/news/claude-2-1}} This makes it possible to input an entire document and perform location tracking as an end-to-end task.

Our pipeline consists of three steps for the map: (1) per-document location-graph extraction; (2) combining all graphs; and (3) visualization, and two steps for the mapping: (1) path extraction for a given document; and (2) visualizing the path on the combined graph. See Figure \ref{fig: figure1} for an overview.

Here we describe the details for each step.

\subsection{Per-document location-graph extraction}

For each document, we first extract a graph of the mentioned locations and their relationships.
We use highly detailed instructions to create a graph of the mentioned locations. The full prompt is provided in Appendix \ref{prompt1}.

Following work that suggests that LLMs have self-correction capabilities \cite{pan2023automaticallycorrectinglargelanguage}, we added a revision step, instructing the model to check if the answer is consistent and return a revised answer. The prompt is provided in Appendix \ref{prompt2b}. 

\subsection{Combining the graphs into a map}

To combine the obtained graphs into one global map, we first need to make sure that each location has only one label.
Once we have one name per node, we can use the name as the identifier and create a graph with the new set of names and with all edges (removing duplicates).

To create a conversion dictionary for double names, we instruct an LLM to combine nodes referring to the same location. The full prompt is in Appendix \ref{prompt3}.
The conversion dictionary is a single output and it allows human proofing. 

After aligning the node names, all nodes and edges are used to create a large map. We apply some simple heuristics to sparsify the edges -- we discard edges between nodes of the same type (e.g., no edge from country to country), and edges that go against the type hierarchy (i.e., we discard edges from country to city or from continent to country). 

\paragraph{Per-testimony trajectory extraction.}
Following an answer about the locations in a document, the model is tasked to generate the trajectory, with the possible locations being the nodes of the previously obtained graph. 
The full prompt is provided in Appendix \ref{prompt2}.
Here too, we ask the model to revise its answer (see Appendix \ref{prompt2b}).

\paragraph{Plotting the maps and trajectories.}
Using the Networkx\footnote{\url{https://networkx.org/}} package for visualization, we plot both the combined graph and single trajectories on it.

\begin{table*}
\centering
\begin{tblr}{
  row{2} = {c},
  cell{1}{1} = {r=2}{},
  cell{1}{2} = {c=3}{c},
  cell{1}{5} = {c=3}{c},
  cell{1}{8} = {r=2}{c},
  cell{3}{2} = {c},
  cell{3}{3} = {c},
  cell{3}{4} = {c},
  cell{3}{5} = {c},
  cell{3}{6} = {c},
  cell{3}{7} = {c},
  cell{3}{8} = {c},
  cell{4}{2} = {c},
  cell{4}{3} = {c},
  cell{4}{4} = {c},
  cell{4}{5} = {c},
  cell{4}{6} = {c},
  cell{4}{7} = {c},
  cell{4}{8} = {c},
  cell{5}{2} = {c},
  cell{5}{3} = {c},
  cell{5}{4} = {c},
  cell{5}{5} = {c},
  cell{5}{6} = {c},
  cell{5}{7} = {c},
  cell{5}{8} = {c},
  cell{6}{2} = {c},
  cell{6}{3} = {c},
  cell{6}{4} = {c},
  cell{6}{5} = {c},
  cell{6}{6} = {c},
  cell{6}{7} = {c},
  cell{6}{8} = {c},
  cell{7}{2} = {c},
  cell{7}{3} = {c},
  cell{7}{4} = {c},
  cell{7}{5} = {c},
  cell{7}{6} = {c},
  cell{7}{7} = {c},
  cell{7}{8} = {c},
  cell{8}{2} = {c},
  cell{8}{3} = {c},
  cell{8}{4} = {c},
  cell{8}{5} = {c},
  cell{8}{6} = {c},
  cell{8}{7} = {c},
  cell{8}{8} = {c},
  cell{9}{2} = {c},
  cell{9}{3} = {c},
  cell{9}{4} = {c},
  cell{9}{5} = {c},
  cell{9}{6} = {c},
  cell{9}{7} = {c},
  cell{9}{8} = {c},
  cell{10}{2} = {c},
  cell{10}{3} = {c},
  cell{10}{4} = {c},
  cell{10}{5} = {c},
  cell{10}{6} = {c},
  cell{10}{7} = {c},
  cell{10}{8} = {c},
  cell{11}{2} = {c},
  cell{11}{3} = {c},
  cell{11}{4} = {c},
  cell{11}{5} = {c},
  cell{11}{6} = {c},
  cell{11}{7} = {c},
  cell{11}{8} = {c},
cell{12}{2} = {c},
  cell{12}{3} = {c},
  cell{12}{4} = {c},
  cell{12}{5} = {c},
  cell{12}{6} = {c},
  cell{12}{7} = {c},
  cell{12}{8} = {c},
  vline{2} = {1-12}{},
  vline{5} = {1-12}{},
  vline{8} = {1-12}{},
}
\hline \hline
\textbf{Model}        & \textbf{{\sc R-Edit}} &      &      & \textbf{{\sc Edit}}   &      &      & \textbf{Length} $\pm$ \textbf{STD}\\
 & \textbf{{\sc SF-Ref}} & \textbf{{\sc Ref1}} & \textbf{{\sc Ref2}} & \textbf{{\sc SF-Ref}} & \textbf{{\sc Ref1}} & \textbf{{\sc Ref2}} &\\
             \hline \hline
{\sc SF-Ref}       & -   &\textbf{ 0.25}    & 0.29 & -   & 2.7    & 2.96 & 10.6 $\pm$ 2.33\\
{\sc Ref1}         & \textbf{0.25}   & -    & 0.5 & 1.41   & -    & 1.13 & 20.17 $\pm$ 5.34\\
{\sc Ref2}         &  0.29  & 0.5 & -    &  2.96  & 1.13 & -    & 36 $\pm$ 6.98\\
\hline
Random        & 1   & 1    & 1 & 1      & 1    &  1   & -\\
Frequent       & 0.709   & 0.709 & 0.93 & 0.79   & 0.79 & 0.93 & -\\
SpaCy       &  0.36  & 0.82 & 0.78 &  15.52  & 8.39 & 4.16 & 158  $\pm$ 20.51\\
\hline
GPT-4o mini  & 0.42   & 0.49 & 0.45 & 1.06   & 1.58 & 3.09&  11.5 $\pm$ 3.77\\
GPT-4o       & \textbf{0.36}   & 0.39  & \textbf{0.39} & 0.93 & 1.64 & 3.46 &  9.66 $\pm$ 2.75\\
o1-mini      & 0.4   & \textbf{0.34} & 0.56 & 0.86   & 1.26 & 2.13 &  11.67 $\pm$ 4.46\\
Llama-3.1-8B &  0.46 & 0.66 & 0.74 &  1.72 & 1. & 1.45 & \textbf{20 $\pm$ 5.7 }
\end{tblr} 
\caption{Edit distances and lengths for the references and models. We report the normalized Edit and recall-focused Edit distances for the different models on all references. We also report the distances between the references. For comparison, we report the distances for random choices and for a constant choice of the most frequent location. We also report the average lengths and standard deviations. {\sc SF-Ref} and {\sc Ref1} contain $6$ testimonies for which scores were computed. {\sc Ref2} contains $3$ testimonies.}
\label{table: 1}
\end{table*} 

\section{Experimental Setup}

\subsection{Implementation Details}

We ran the pipeline on a set of $402$ testimonies and on a set of $75$ Lake District works. We used the texts only, without any labels. We made minor changes in the prompts to fit the Lake District domain.

We used mainly {\it GPT-4o mini} which has a context length of $128K$ tokens.\footnote{\url{https://openai.com/index/gpt-4o-mini-advancing-cost-efficient-intelligence/}} The price for running the pipeline on the entire testimony set was $\approx 7$ \$. 

We used revision steps in the per-document parts.
For the name-conversion dictionary, we used GPT-4o.
We found that this step required manual proofing of the resulting list, adding some merges.

\subsection{Evaluation}

\paragraph{Trajectories.}
With reference trajectories (\S \ref{sec:eval}), we can evaluate the output with versions of the edit distance (\S \ref{trajectory_eval}). However, we must ensure that locations in the reference and prediction are given in the same format.
For this, we used GPT-4o with instructions to align locations from the predicted sequence to locations in the reference sequence. The full prompt is provided in Appendix \ref{prompt4}.\footnote{Although the model was not told how the predictions were generated, it is possible that this process introduces biases and noise. Manual inspection did not reveal such trends.}

We evaluated three OpenAI models: gpt-4o-mini-2024-07-18 (\textit{GPT-4o mini}), gpt-4o-2024-05-13 (\textit{GPT-4o}), and\textit{o1-mini},\footnote{\url{https://platform.openai.com/docs/models}} and the open-source model Llama-3.1-8b.\footnote{\url{https://ai.meta.com/blog/meta-llama-3-1/}} All tested models accept an input context of $128K$ tokens.

For comparison, we measured the distances on three simple methods. The first is independent random guessing of the length of the reference trajectory. We randomly selected from the set of nodes in the combined graph. The second is a fixed choice of the most common location in the reference trajectory. We note that the second method receives additional data that the other methods are not given.
The third method is based on SpaCy's NER model.\footnote{\url{https://spacy.io/api/entityrecognizer}. We used the medium-size English model.} The trajectory is simply the sequence of GPE and LOCATION entities output by the model, omitting consecutive identical entities.

\paragraph{Maps.}
When provided with a reference map, we can evaluate the output map with accuracy metrics (\S \ref{map_eval}). Using the reference map for the Lake District works (\S \ref{sec:eval2}), we evaluate the output map of the pipeline when run on the CLDW.

For comparison, we created a random tree with the same locations. This tree follows the same principles of the reference map (\S \ref{sec:eval2}), but the connections between the levels (e.g., to what city we connect a natural place) are random.

%%%%%%%%%%%%%%%%%%%%%%%%%%%%%%%%
\section{Results}

Here we report the partial evaluation results and present the statistics of the outputs and some examples of the resulting maps and paths.

\paragraph{Testimonies.}
We ran the pipeline on $402$ testimonies from the SF. 
The resulting graph has $2533$ nodes and $2785$ edges. 
The median length of the trajectories is $12$.

In Figure \ref{fig: figure1} we present a view of the map and a trajectory on it. This is the trajectory of a survivor that started in Czechoslovakia, went through the Theresienstadt Ghetto and Auschwitz, and ended up in New Zealand. 
Figure \ref{fig: figure2} shows an enlarged example snippet around Israel.

In Table \ref{table: 1} we report normalized Edit distances (compared to the reference data), including those for the modified version.  
We also report the trajectory length for the different models. 

\begin{table}
\centering
\begin{tabular}{l|ccc}
\hline\hline
\textbf{Model} & \textbf{Precision }& \textbf{Recall} & \textbf{F1} \\ \hline
{\sc Random Tree} & 0.09 & 0.09 & 0.09 \\ 
GPT-4o-mini & \textbf{0.23} & \textbf{0.22} & \textbf{0.23} \\ 
\end{tabular}
\caption{Accuracy scores when compared to the reference map. We report precision, recall, and F1-score for the map produced by our pipeline and for a tree with random connections between levels.}
\label{table: 2}
\end{table}

\paragraph{Lake District.}
We ran the pipeline on $75$ Lake District works.
The resulting graph has $783$ nodes and $863$ edges. 
The median trajectory length is $19$.

In Table \ref{table: 2} we report the accuracy scores of the resulting map when compared to the reference map of the Lake District (\S\ref{sec:eval2}).\footnote{Since the reference map was constructed with no edges between nodes of the same type, for a fair comparison, we modified the output map to fit this format. For example, in any case of a natural location that is a child of another natural location, we removed the connecting edge and connected both natural locations to the common city. Additionally, we removed from both maps all the nodes that did not have GIS labels.}

In Appendix \ref{appendix: B} we provide plots of the map extracted by the pipeline for the Lake District works.

\begin{figure*}[htbp]
    \centering
    \fontsize{7.3pt}{8pt}\selectfont
    \begin{subfigure}{1\textwidth}
        \centering
        \begin{tabular}{c|llllllllllll}
        \textbf{ID} & \multicolumn{12}{c}{\textbf{Trajectory}} \\ \hline \hline
        37250 & Bratislava & \begin{tabular}[c]{@{}l@{}}Czecho-\\ slovakia\end{tabular} & ... & \multicolumn{1}{c}{---} & Budapest & Brooklyn & USA & Auschwitz & ... & Lüneburg & Germany & Łódź \\
        29464 & Chust & \begin{tabular}[c]{@{}l@{}}Czecho-\\ slovakia\end{tabular} & ... & Romania & Budapest & \multicolumn{1}{c}{---} & \multicolumn{1}{c}{---} & Auschwitz & ... & \multicolumn{1}{c}{---} & Germany &  \\ 
        \end{tabular}
        \caption{Trajectories and alignment for testimonies 37250 and 29464}
        \label{tab:1}
    \end{subfigure}

    \vspace{0.1cm}
    \begin{subfigure}{1\textwidth}
        \centering
        \begin{tabular}{c|lllllllll}
        
        \textbf{ID} & \multicolumn{9}{c}{\textbf{Trajectory}} \\ \hline \hline
        28857 & Krakow & \begin{tabular}[c]{@{}l@{}}Krakow\\ Ghetto\end{tabular} & Plaszow & Auschwitz & Brunnlitz & Long Beach & New York & \multicolumn{1}{c}{---} & USA \\
        28872 & Prague & Terezin & \multicolumn{1}{c}{---} & Auschwitz & \begin{tabular}[c]{@{}l@{}}Christian-\\ stadt\end{tabular} & Kladno & Havertown & \begin{tabular}[c]{@{}l@{}}Pennsyl-\\ vania\end{tabular} & USA \\
        
        \end{tabular}
        \caption{Trajectories and alignment for testimonies 28857 and 28872}
        \label{tab:1}
    \end{subfigure}
    \caption{Examples of similar trajectories. The distance was measured with a modified Edit distance that takes into account the similarity between locations.}
    \label{fig:tables}

    % \end{tiny}
\end{figure*}

\section{Discussion}

The results for the map show that the pipeline can produce meaningful maps. Quantitative results for trajectories show that all models are substantially better than the baselines in terms of the recall-focused metric. GPT-4o is the best-performing model (according to these metrics), while all LLMs are comparable to the agreement between humans.

We find that trajectory lengths vary substantially between different sources. We can also see that the Edit distance is highly influenced by the lengths of the predictions (as evidenced in the baselines).

Altogether our experiments demonstrate that LLMs are capable, with appropriate prompting, of producing maps from large corpora and representing trajectories over the maps. This trajectory can be used as an abstract representation of the document, which can be useful for downstream applications, such as story understanding. We give here two examples use cases from the domain of Holocaust studies. 

\begin{enumerate}[leftmargin=*,topsep=0pt,itemsep=-1ex,partopsep=1ex,parsep=1ex]
    \item 
\textbf{Trajectory Similarity and Alignment:}
For a pair of locations, we can define meaningful similarity measures based on the graph. For example, we use the (undirected) distance on the graph (so, for example, two towns in Poland will be closer to each other than to a city in the USA). In addition, since we extracted the types of locations, we can put special emphasis on Holocaust-specific locations (like ghettos and camps). We can define a distance that penalizes type mismatches.

Provided with a point-wise distance measure (i.e., the distance between two locations) we can derive a trajectory-wise distance. 
For example, we can use versions of the edit distance or Dynamic Time Warping \cite{vintsyuk1968speech} built upon the point-wise distance. This type of measure has the benefit of generating an optimal alignment between the trajectories, which in itself can be highly beneficial.
A similarity measure allows us to perform unsupervised analysis such as clustering or outlier detection. 
In Figure \ref{fig:tables} we provide examples of testimony pairs that were close in terms of a modified Edit distance (both in the top 5). The modification was in the substitution, where the price of substitution was proportional to the distance on the graph and penalized by mismatching location types.

\item
\textbf{Local Alignment:}
Alignment can also be performed locally by looking at specific pairs (i.e., one transition) or triplets (i.e., two consecutive transitions). As our pipeline links locations with specific parts of the testimonies, we can use common transitions to extract and analyze corresponding parts in different testimonies.
In Table \ref{tab:pairs} we report the most common Holocaust-related transitions.\footnote{We omit non-Holocaust-related transitions as the most common ones are trivial, e.g., Brooklyn to New York.} We report only transitions that appear in at least $4$ different testimonies (out of $402$).
\end{enumerate}

\begin{table}
\begin{small}
    \centering
    \begin{tabular}{ll|c}
         \multicolumn{1}{c}{\textbf{From}} & \multicolumn{1}{c}{\textbf{To}} & \textbf{Count} \\
         \hline \hline
         Auschwitz & Birkenau  & 20 \\
         Theresienstadt & Auschwitz  & 18\\
         Auschwitz & Bergen-Belsen & 10\\
         Budapest & Auschwitz & 7 \\
         Birkenau & Auschwitz & 5\\
         Auschwitz & Mauthausen & 5\\
         Auschwitz & Theresienstadt & 5\\
         Mauthausen & Gunskirchen &  4\\
         Dachau & Auschwitz & 4\\
         Plaszow & Auschwitz & 4\\
    \end{tabular}
    \caption{Most common Holocaust-related transitions. The count is the number of occurrences in the set of $402$ testimonies.}
    \label{tab:pairs}
\end{small}
\end{table}

\paragraph{Challenges.}
% We note that our model faces some challenges that might require further research.
A general challenge encountered is the ambiguity of the unsupervised task. Different sources gave different trajectory lengths, despite our efforts to create clear guidelines. This seems to be due to unavoidable disagreements of what locations should be included (e.g., how significant is it to the story and what level of details should be included). While the additional guidelines led to trajectories closer to the {\sc SF-Ref} trajectories, there are still substantial differences between sources. We discuss this in detail in Appendix \ref{app:effect}.

A practical challenge we face is the ambiguity of location names. Many places appear with partial names, shared with other places. In some cases, disambiguation can be done with information from the document itself, but in some cases, the information is lacking altogether.

Another challenge is political changes. The testimonies span entire life stories, in which the political status of many countries changed. An example of this is the Theresienstadt Ghetto (with the occupation of Czechoslovakia by Germany). This can be seen in our example map (Figure \ref{fig: figure1}, where many countries in Europe are intertwined through places that are attributed to more than one country.

\section{Conclusion}
\vspace{-.08cm}

We presented and defined the task of unsupervised trajectory extraction. We built and demonstrated a pipeline for the task, based on GPT-4o-mini. Our demonstration shows that new models are capable of extracting meaningful trajectories from full testimonies, without breaking them into segments. We also showed some use cases for further research. 

Our work demonstrates the role of LLMs as a valuable tool for the Humanities \cite{aguiar2024final}, and specifically for Holocaust research. NLP technology has recently been applied to the analysis of Holocaust testimonies \citep{artstein-etal-2016-new, wagner-etal-2022-topical}. By leveraging NLP, researchers can extract valuable insights from the vast array of testimonies (comprising tens of thousands), instead of limiting themselves to small-scale studies. 

\section*{Ethical Considerations}

We followed the guidelines given by the SF archive. Although the testimonies were not given anonymously, no identifying details are included in our analysis. Our codebase and scripts will be released, but they do not contain any data from the archives. Permission to use the data and trained models used in our work for research purposes requires approval from the SF archive.

\section*{Limitations}

Since our task generates sequences of locations without a taxonomy of possible locations, it has a range of possible outputs. 
This makes the evaluation challenging. It is difficult to determine when the distance from the reference is due to poor performance and when it is due to other possible outputs.

Also, the evaluation method itself is limited -- it focuses on the recall of reference locations and not on the precision of the predicted ones. It also involves LLMs in the process of aligning the location descriptions, which can lead to mistakes or biases.

The combination of the locations into a unified graph turned out to be challenging and required human intervention. However, this intervention is done only once and does not require reading the testimonies.

The evaluation of the graph also has limitations. The reference graph is heuristically constructed and may have incorrect connections. It is constructed only with nodes that were output by the model since we do not require all named entities to be part of the map. Therefore, this evaluation is precision focused. 

\section*{Acknowledgments} 
The authors acknowledge the USC Shoah Foundation - The Institute for Visual History and Education for its support of this research.
This research was supported by grants from the Israeli Ministry of Science and Technology and the Council for Higher Education, the Alfred Landecker Foundation, and the Federmann cyber security research center.

% \nocite{*}
\bibliography{anthology,custom}
% \bibliographystyle{acl_natbib}
% \bibliography{anthology}

% \newpage
\appendix
\section{Prompts for the Pipeline} \label{appendix: A}

Here we provide the prompts that were used for the models. These prompts contain detailed instructions that were established in coordination with human annotation. 

\subsection{Per-testimony location-graph extraction} \label{prompt1}
The prompt was the following:
\begin{quote}
    I'll give you a Holocaust testimony. \\
    I want you to give me a JSON representing the graph of the mentioned locations (proper and common) and any known relations between them. Locations can be GPEs (like country or city) or significant facilities (like army camps, ghettos, concentration camps and death camps).  \\
    % Some important points:\\
    % 1. Make sure the nodes contain locations only and not anything else (no nodes for events or people). \\
    % 2. Give the nodes a type based on the type of location. The types should include: City, Country, Village, Ghetto, Army Camp, Concentration Camp, and Death Camp. \\
    % 3. Keep the graph as full as possible,  so, for example, if a place in a city in country is mentioned, there should be nodes for the place, the city, and the country. Separate a district from a city description into two nodes.\\
    % 4. The graph should include relations between locations (i.e., A is in B).  Make sure that the direction of an edge is that of inclusion if relevant (that is, if A is in B then the edge should be from A to B).\\
    % 5. Make sure to avoid double entries. \\
    % 6. Give me the graph as JSON dictionary, with the "nodes" field indicating a list of nodes,  and "edges" indicating a list of edges. These nodes and edges should be in a format that can create a python networkx graph. Make sure the nodes are given as a list of tuples, in which the first value is the name and the second is a dictionary with the type (as described above) The edges should be in a list of tuples, each containing two names (see example). \\
    Some important points:\\
    1. Make sure the nodes contain locations only and not anything else (no nodes for events or people).\\
    2. Give the nodes a type based on the type of location. The types should include: City, Country, Village, Ghetto, Army Camp, Concentration Camp, and Death Camp. Do not mark exact addresses.\\
    3. Facilities should be included if they’re significant (in terms of events happening there). For example, being near a police station is not significant, but if there’s a significant story going on inside then it should be marked.\\
    4. Unknown cities/towns/villages should be marked (e.g. for a town near Cracow, mark “Town near Cracow”). In the "map" part these should be connected to the reference point (e.g., “Cracow”), if there is one, or to Poland itself. The same for cases like a forest near some place.\\
    5. Also hiding places can be marked as a place (that is, use “Hiding place near …”). In the “map” part it will be connected to the close by city or facility.\\
    6. The forest should be mentioned if the witness stays or hides there. Just going through (without much else happening) can be omitted.\\
    7. Keep the graph as full as possible,  so, for example, if a place in a city in a country is mentioned, there should be nodes for the place, the city, and the country. Separate a district from a city description into two nodes.\\
    8. The graph should include relations between locations (i.e., A is in B).  Make sure that the direction of an edge is that of inclusion if relevant (that is, if A is in B then the edge should be from A to B). The relation is either inclusion (i.e., city A in country B) or proximity (i.e., city A near city B).\\
    9. Every location should be connected (directly or indirectly) to a country.\\
    10. Make sure to avoid double entries. \\
    11. Give me the graph as JSON dictionary, with the "nodes" field indicating a list of nodes,  and "edges" indicating a list of edges. These nodes and edges should be in a format that can create a python networkx graph. Make sure the nodes are given as a list of tuples, in which the first value is the name and the second is a dictionary with the type (as described above) The edges should be in a list of tuples, each containing two names (see example). 
    
    Here is an example (from a different testimony): \\
    ```json\\
    {\\
    "nodes": <Here we provide an example list of locations>,\\
    "edges": <Here we provide an example relations between the locations>\\
    }\\
    ```
    
    This should all be based on the text. 
    
    Testimony:
    <Here we add the testimony divided into numbered segments>
    
\end{quote}

\subsection{Per-testimony trajectory extraction} \label{prompt2} 
The trajectory was extracted with the following prompt:

\begin{quote}
    Now, can you give a graph with the trajectory of the witness' movements? That is, give a list of locations where he is. 
    All location nodes should be nodes from the networkx graph you gave before. The nodes should have a field noting the sentence number in the text in which the witness was in that location.\\
    The edges should be between each adjacent node by order of the testimony. \\

    Some important points:\\
    1. Include all of the places in the testimony (also the ones after the war), as long as the interviewee is there himself/herself, and a description of events relating to the place is given.\\
    2. Only include places where the interviewee is staying/traveling to, not if only relating to family/friends. Do mark a place if the mention implies that the interviewee went there too (e.g., “my father got a job in Berlin, where we rented a small apartment”).\\
    3. Mark each stay or travel to a place only once. If the story repeats a specific stay that has already been annotated, there is no need to mark it again. Different travels, even if they are to the same place, should all be marked separately.\\
    4. Journeys/travels should be marked even if no specific named place is mentioned, as long as there is a significant story, (e.g. trek through Europe, sea voyage).\\
    5. List the place of birth (and not the place of interview) at the beginning and the place of the interview at the end.\\
    6. If it is clear that a specific place includes a significant story (even if the story is not being told), it should be marked (e.g., a journey through the Alps).\\
    7. General customs and traditions of a specific place, or general experiences (e.g.: "In Poland, if you didn't pass first grade they keep you another year." or "We experienced antisemitism in Poland"): Only mark it if no place connected to it is annotated yet and the interviewee was really staying in that place. If for example "Cracow" is marked as a location in the interview, and the interviewee mentions Polish customs/experiences, there is no need to mark "Poland" separately.\\
    8. Give me a graph in JSON format (like in the example). The response should be a valid JSON only, without comments or additional text.
    
    % For each edge, add the method of transportation that can be inferred from the text. Methods include: By foot, By car, By train, and By plane. If the method is unknown give Unknown.\\
    % Give me a graph in JSON format (like in the example). \\
    
    For example:\\
    ```json \\
    "nodes": <Here we provide an example list of locations with their place in the testimony>,\\
    "edges": <Here we provide an example relations between adjacent locations, with the method of transportation>\\
    ```
\end{quote}

\subsection{Revision prompts} \label{prompt2b} 
Revision for the graph was done with the following prompt:

\begin{quote}
    Go over your answer and make sure that it is consistent. Check the types of the nodes and the direction of the edges. Make sure that the nodes are locations only and that there are no double entries.
    Give your (possibly) corrected answer in the same JSON format.
\end{quote}

Revision for the trajectory was done with the following prompt:

\begin{quote}
    Go over your answer and make sure that it is consistent. 
    Make sure that: (1) the sentence numbers are in ascending order; (2) a node does not repeat without other nodes between; 
    (3) there are edges between adjacent nodes; (4) a long description of a location is not repeated as a separate node (e.g., "Brooklyn, New York" should be one node and not two).
    
    Give your (possibly) corrected answer in the same JSON format.
\end{quote}

\subsection{Combining the graphs into a map} \label{prompt3}
For combining the locations, we use the following prompt:
\begin{quote}
    I'll give you (in JSON format) a list of place names. I want you to see if there are any places that appear twice but with different names. \\
    Give me a JSON with a list of lists, where the inner list is the multiple names that describe the same place (and both appear in the input). No need to return unique names (i.e., lists with one element). \\
    Convert names only if you are positive that they are the same, e.g., different spellings or a longer description of the same place (like US, USA, America etc.). \\
    Make sure to maintain the exact spelling that appeared, including special characters.
    Make sure to give only the JSON format with no additional text.\\
    
    For example, if the input is:\\
    ```json \\
    <Here come some examples of lists of names describing the same place>
    ```\\
    
    Here is the input:\\
    <Here comes a sorted list of the locations>
\end{quote}

\subsection{Evaluation} \label{prompt4}
For aligning locations from the predicted sequence to locations in the reference sequence, we use the following prompt:

\begin{quote}
    I have a list of predicted locations and a list of locations from the gold standard.
    For each location in the predicted list, I want you to find a corresponding location in the gold standard list if it exists (even if it's written differently). 
    In the case it exists, give me the id of the corresponding location in the gold standard list. If it doesn't exist, give me -1.
    
    Here is an example:\\
    For predicted locations: ["Warsaw (Ghetto)", "Luck", "Warsaw", "New York"],\\
    and gold-standard locations: ["Lutsk", "The Warsaw ghetto"]]
    
    The output should be the JSON: \\
    {"ids": [1, 0, -1, -1]}
    
    Make sure to follow the instructions and give the output in the correct format.
    
    Predicted locations: <predicted path>, \\
    Gold-standard locations: <reference path> 
\end{quote}

\section{Effect of Detailed Instructions} \label{app:effect}
The reported results were obtained with annotations and LLM inference with detailed instructions.
In initial experiments, the tasks (for humans and LLMs) were conducted with loose guidelines.
Specifically, points $3-9$ in Prompt \ref{prompt1} and points $1-7$ in Prompt \ref{prompt2} did not appear in the initial experiments.

One annotator performed annotation both with and without detailed guidelines. We report the average trajectory length and Edit distances for this annotator as well as the LMs in Table \ref{tab:lengths}.

Regarding the trajectory lengths, the clear trend is that the ranking between the models is preserved, however the trajectories are all shorter.
Regarding the Edit distances, we see a clear improvement with respect to the {\sc SF-Ref}. 
Nevertheless, the variance between annotators and between models is still high, suggesting that it is hard to design strict guidelines for this task.

\begin{table*}
    \centering
    \begin{tabular}{ll|c|cc}
    \hline \hline
         \textbf{Model} & \textbf{Details?}& \textbf{Length} & \textbf{{\sc Edit}}& \textbf{{\sc R-Edit}} \\
         \hline \hline
         {\sc Ref1} & No & 27.8 $\pm$ 4.45 & 0.29 & 2.7 \\
          & Yes & 20.17 $\pm$ 5.34 & 0.25 & 1.41\\
         \hline \hline
        GPT-4o mini  & No & 14 $\pm$ 4.6 & 0.51 & 1.21\\
          & Yes & 11.5 $\pm$ 3.77 & 0.42 & 1.06\\
          \hline
        GPT-4o       & No & 11.6 $\pm$ 4.32 & 0.39 & 0.85 \\
         & Yes &  9.66 $\pm$ 2.75 & 0.36 & 0.93\\
         \hline
        o1-mini    & No & 15.2 $\pm$ 3.06 & 0.57 & 1.32 \\
         & No & 11.67 $\pm$ 4.46 & 0.4 & 0.86 \\
         \hline
        Llama-3.1-8B & No & 22 $\pm$ 7.48  & 0.55 & 1.68\\
         & Yes  & 20 $\pm$ 5.7  & 0.46 & 1.72\\
         \hline 
    \end{tabular}
    \caption{Trajectory lengths and Edit distances (from the {\sc SF-Ref}) with and without detailed instructions.}
    \label{tab:lengths}
\end{table*}

\section{Lake District Maps} \label{appendix: B}
\begin{figure*}[t]
\centering
\includegraphics[scale=0.58]{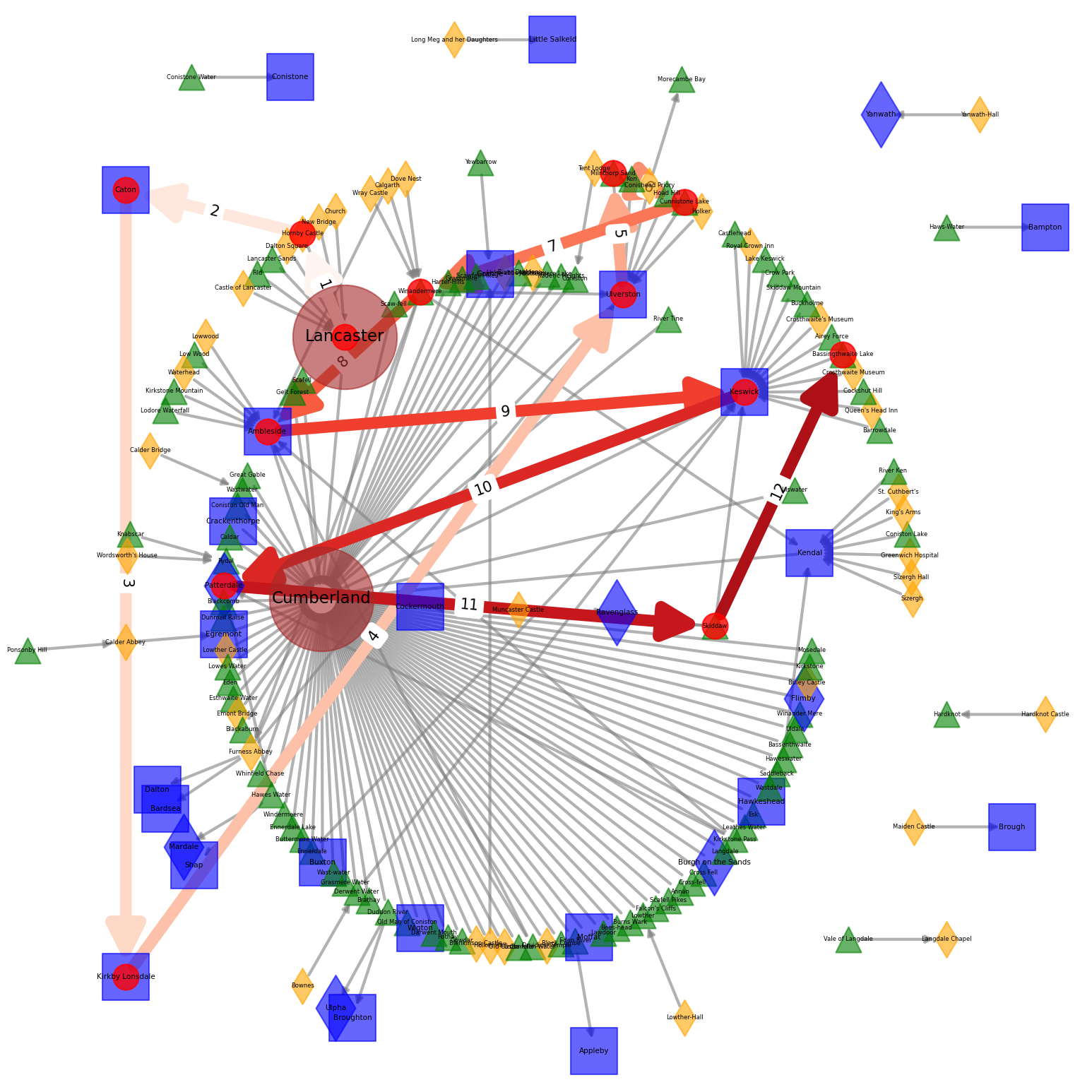}
\caption{Visualization of a location map with a single trajectory. The map was generated based on 75 works with GPT-4o-mini. Some low-degree nodes were removed for clarity. Counties are displayed as brown circles, with the size depending on the degree. Cities and Villages are blue squares. Natural locations are green triangles and Facilities are yellow diamonds. The trajectory is in shades of red, getting darker with the progression of the trajectory.} \label{fig: figure5}
\end{figure*}

\begin{figure}[t]
\centering
\includegraphics{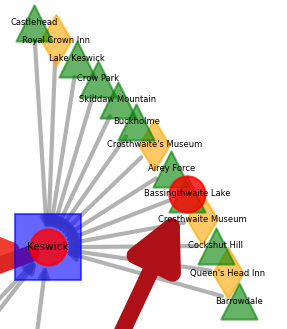}
\caption{Snippet from the map that includes Keswick and locations within it.} \label{fig: figure6}
\end{figure}

Here we provide some examples from the output for the CLDW.
In Figure \ref{fig: figure5} we plot the resulting map and in Figure \ref{fig: figure6} we provide a snippet from it.

\end{document}